\documentclass[journal]{IEEEtran}

\ifCLASSINFOpdf
\else
   \usepackage[dvips]{graphicx}
\fi
\usepackage{url}
\usepackage[table,xcdraw]{xcolor}
\usepackage{graphicx}
\usepackage{amsmath}

\def\I{\mathcal{I}}
\def\D{\mathcal{D}}
\def\P{\mathcal{P}}
\def\X{\mathcal{X}}
\def\Y{\mathcal{Y}}
\def\S{\mathcal{S}}
\def\R{\mathrm{R}}
\def\p{\mathbf{p}}
\def\x{\mathbf{x}}
\def\y{\mathbf{y}}

\begin{document}

\title{Point Cloud Quality Assessment using 3D Saliency Maps}

\author{Zhengyu Wang, Yujie Zhang, Qi Yang, Yiling Xu, \IEEEmembership{Member, IEEE}, Jun Sun, and Shan Liu

\thanks{This paper is supported in part by National Natural Science Foundation of China (61971282, U20A20185). The corresponding author is Yiling Xu(e-mail: yl.xu@sjtu.edu.cn).}
\thanks{Zhengyu Wang, Yujie Zhang, Yiling Xu and Jun Sun are with the Cooperative Medianet Innovation Center, Shanghai Jiao Tong University, Shanghai 200240, China, (e-mail: wang\_zheng\_yu@sjtu.edu.cn; yujie19981026@sjtu.edu.cn; yl.xu@sjtu.edu.cn; junsun@sjtu.edu.cn)}
\thanks{Qi Yang is with Tencent MediaLab, Shanghai, China. (e-mail: chinoyang@tencent.com)}
\thanks{Shan Liu is with Tencent MediaLab, Palo Alto, United States. (e-mail: chinoyang@tencent.com; shanl@tencent.com)}

}

\markboth{Journal of \LaTeX\ Class Files, Vol. 14, No. 8, August 2015}
{Shell \MakeLowercase{\textit{et al.}}: Bare Demo of IEEEtran.cls for IEEE Journals}
\maketitle

\begin{abstract}
Point cloud quality assessment (PCQA) has become an appealing research field in recent days. Considering the importance of saliency detection in quality assessment, we propose an effective full-reference PCQA metric which makes the first attempt to utilize the saliency information to facilitate quality prediction, called point cloud quality assessment using 3D saliency maps (PQSM). Specifically, we first propose a projection-based point cloud saliency map generation method, in which depth information is introduced to better reflect the geometric characteristics of point clouds. Then, we construct point cloud local neighborhoods to derive three structural descriptors to indicate the geometry, color and saliency discrepancies. Finally, a saliency-based pooling strategy is proposed to generate the final quality score. Extensive experiments are performed on four independent PCQA databases. The results demonstrate that the proposed PQSM shows competitive performances compared to multiple state-of-the-art PCQA metrics.
\end{abstract}

\begin{IEEEkeywords}
objective quality assessment, point cloud, saliency map, structural similarity
\end{IEEEkeywords}

\IEEEpeerreviewmaketitle

\section{Introduction}

\IEEEPARstart{W}{ith} the rapid development of 3D capturing technologies, point clouds (PCs) have emerged as a popular and prominent format to represent 3D photo-realistic content. A PC is composed of a great number of disordered points, and each point is attached with the information of its 3D coordinates and additional attributes such as RGB color and normals. In recent years, we have witnessed the occurrence of many applications based on PCs (e.g., navigation \cite{Navigation}, industrial robotics\cite{robotics} and mixed reality\cite{mixedreal}, etc) which require PC data with high quality. However, a variety of distortions could be induced during the relevant processing of PCs, such as acquisition, compression \cite{Compre1}\cite{Compre2}, transmission \cite{transmiss}, and rendering, which may impair the perceptual quality of the human visual system (HVS). Therefore, point cloud quality assessment (PCQA) has become an appealing research field to ensure provide high-quality human perception service.

PCQA aims to directly reflect or approximate the perception quality of HVS for specific PCs. Subjective PCQA is the most straightforward and reliable method, but also time-consuming and expensive. Thus, effective objective PCQA metrics which correlate well with HVS for practical application are highly desired. In this paper, we focus on the full-reference (FR) objective PCQA metrics. Existing FR-PCQA metrics either project 3D PCs onto 2D planes, utilizing the well-developed image quality assessment (IQA) algorithms; or inferring the perceptual quality by modeling the 3D properties. In terms of the former, projection-based metrics like \cite{SJTUPCQA} project PCs onto several perpendicular image planes and aggregate image-based features to evaluate the PC quality. In terms of the latter, point-wise metrics, such as the point-to-point\cite{p2po} and the point-to-plane\cite{p2pl}, calculate the differences of the Euclidean distance related features between the corresponding point pairs. Meanwhile, some other metrics, e.g., PointSSIM\cite{pointSSIM}, GraphSIM\cite{GraphSIM}, MS-GraphSIM\cite{MSGraphSIM} and MPED\cite{MPED}, construct local neighborhoods to involve the structural features, which usually have better performances due to the fact that HVS is more sensitive to structural information.


One of the things that above methods overlooked is that they hardly consider the vital role of visual saliency in quality prediction. Specifically, the visual saliency is an important characteristic of HVS which can  reflect the human attention distribution.  In the domain of image quality assessment, some metrics integrated with saliency information have achieved great success, such as VSI \cite{VSI}. One reason that there are few PCQA metrics invloved with visual saliency is due to the lag of PC saliency detection.  For the PC saliency detection, most methods are directly conducted on uncolored 3D PCs\cite{SalGuo}\cite{SalDing}, which is usually utilized in machine vision tasks rather than human vision tasks. 



\begin{figure*}[htbp]\label{fig:framework}
\centerline{\includegraphics[width=181mm]{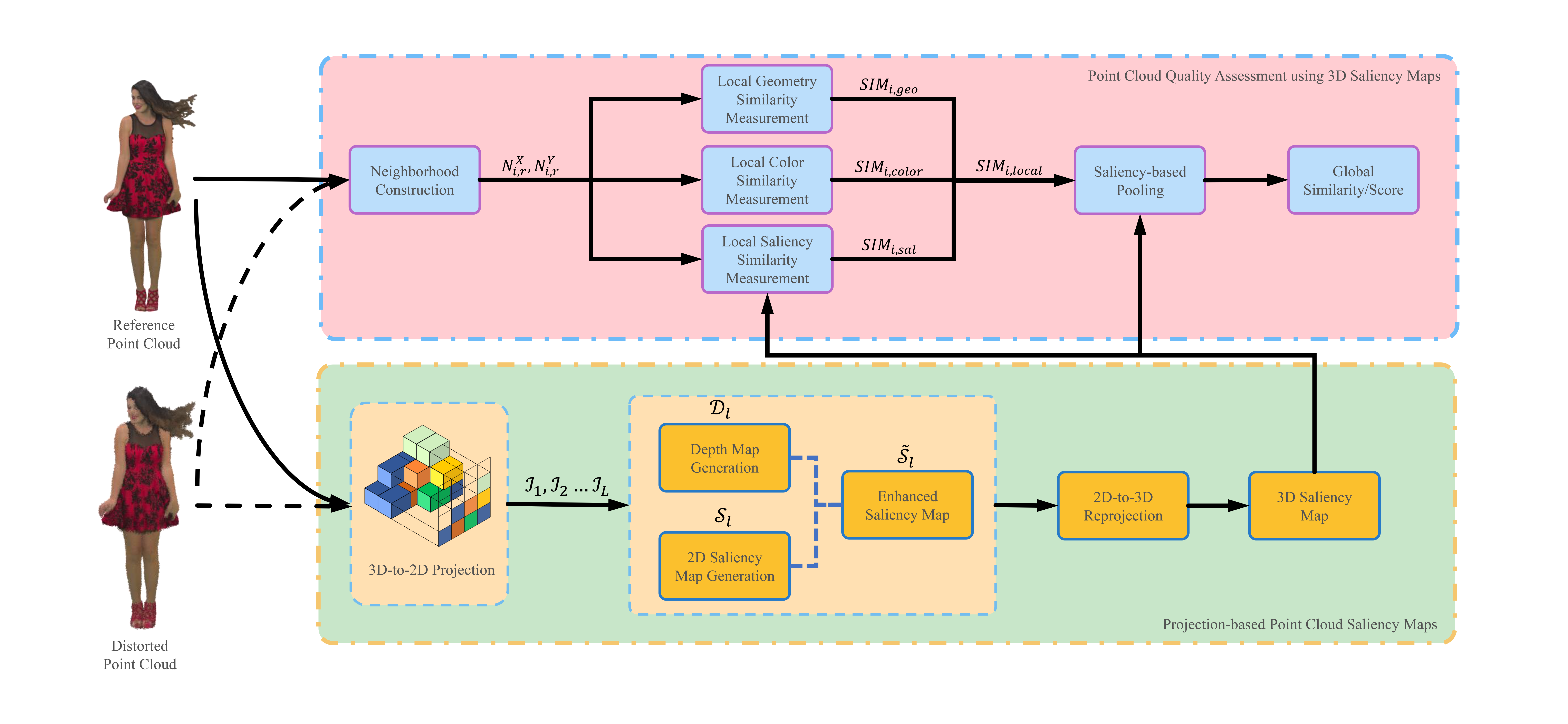}}
\caption{The overall architecture of our proposed method. The box above in pink illustrates the whole PQSM including the neighborhood construction, similarity measurement and pooling strategy. The box below in green illustrates the 3D saliency maps generation including the 3D-to-2D projection, depth maps and 2D saliency maps generation, and 2D-to-3D re-projection.}
\end{figure*}

Inspired by the strategy proposed in \cite{SalMaps}, we use the projection to transfer the achievement of image saliency detection to PC and propose a novel saliency guided metric, called the point cloud quality assessment using 3D saliency maps (PQSM). To our best knowledge, this is the first FR-PCQA metric utilizing the saliency information. Specifically, we first propose to incorporate depth-related maps into the projection-based saliency detection model to involve the intrinsic geometric information of PCs. We then design three descriptors to quantitively measure the distortions on geometry, color and saliency maps. After that, a saliency-based pooling strategy is adopted to generate the final score. Our method is examined by extensive experiments on four publicly accessible PCQA databases. The results show the competitive performance of our proposed model in inferring the perceived quality of distorted PCs.

\section{Method}
In this section, we first illustrate how to enhance the projection-based saliency maps via depth information. Then, we detail the construction of the proposed PQSM. The overall framework of the proposed metric is shown in Fig \ref{fig:framework}. 

\subsection{Projection-based Point Cloud Saliency Maps}\label{sec:projection saliency}
The projection-based saliency generation algorithms for PCs first generate the 2D saliency maps based on projected images, then map these saliency values from pixels in the images to the corresponding points of the original PCs \cite{SalMaps}. Although such methods have achieved considerable correlations with human attention, they ignore the intrinsic geometric characteristic of PCs. Intuitively, in 3D space, the closer an item is to us, the more likely it is to grab our attention \cite{SJTUPCQA}. Therefore, the depth information obtained from the projection process can be leveraged to involve geometry characteristics of PCs in the generation of saliency maps \cite{DepthMatters}.  To simulate the above perceptual property, we utilize depth-related weights to enhance projection-based saliency maps.

Consider one point cloud $\X$ as   $\P=\{\p_1,\p_2, \cdots, \p_N\}\in\R^{N\times6}$, where each point $\p_i=[\p_i^O,\p_i^C]=[x,y,z,R,G,B]$ is a 6-dimension vector containing 3D coordinates $\p_i^O$  and RGB attributes $\p_i^C$ . Let $\P$ be orthogonally projected onto $L$ different 2D planes, on which we obtain $L$ projected texture maps and $L$ depth maps. Note that when deriving depth maps the distance between each projected plane and the closest point in the point cloud is set as 10. We define the $l$-th projected texture map as $\I_l\in\R^{N_l\times3}$ and its corresponding depth map as $\D_l=\{d_{i,l}\}_{i=1}^{N_l}\in\R^{N_l}$, where $N_l$ denotes the pixel number of $\I_l$ or $\D_l$ and $d_{i,l}$ denotes the depth value of $i$-th pixel in $\D_l$. Since one projected image may not contain all the points in $\P$, we have $N_l\le N$. For each texture map, we first calculate its 2D saliency maps based on the established image saliency detection algorithm from \cite{UHF}. For the texture map $\I_l\in\R$, we represent its saliency map as $\S_l=\{s_{i,l}\}_{i=1}^{N_l}\in\R^{N_l}$, where $s_i$ represents the saliency value of $i$-th pixel.

Then, we compute the depth-related weights based on the depth map $\D_l$ and then merge the weights into the 2D saliency map. Specifically, for $i$-th pixel located in the $D_l$, we can derive one new depth enhanced saliency map $\tilde{\S_l}=\{\tilde{s}_{i,l}\}_{i=1}^{N_l}$ as follows,
\begin{equation}\label{eq:depth saliency}
    \tilde{s}_{i,l}= \frac{e^{-\frac{d_{i,l}}{\sigma_s}}}{\sum_{j=1}^{N_l}e^{-\frac{d_{j,l}}{\sigma_s}}}\cdot s_{i,l}  ,
\end{equation}
where $\sigma_s$ is a constant to be determined. According to Eq. (\ref{eq:depth saliency}), we endow higher weights for points with small depths because these points are considered to be closer to human eyes and therefore are more salient.

Finally, for each point in point clouds, we average the depth-enhanced 2D saliency values of its corresponding pixels in $L$ projected images to derive the 3D saliency map $\S_{\P}=\{\p_i^S\}_{i=1}^{N}\in\R^N$, which is utilized for subsequent PCQA task.

\subsection{Point Cloud Quality Assessment using 3D Saliency Maps}
\subsubsection{ Neighborhood Construction}
Consider the reference point cloud as $\X=\{\x_i\}_{i=1}^N\in\R^{N\times6}$ and the distorted point cloud as $\Y=\{\y_j\}_{j=1}^M\in\R^{M\times6}$.  We first define a series of local patch pairs by ball query. Specifically, for a given point $\x_{i}$ in the reference PC, we have two local patches in $\X$ and $\Y$ as,
\begin{equation}
    \begin{aligned}
     N^{\X}_{\x_i,r}&=\{\x_j|\|\x_j-\x_i\|_2\leq r\}, \\
     N^{\Y}_{\x_i,r}&=\{\y_j|\|\y_j-\x_i\|_2\leq r\}, 
    \end{aligned}
\end{equation}
where $r$ denotes the searching radius. We then compute the features over $N^{\X}_{\x_i,r}$ and $N^{\Y}_{\x_i,r}$.

\subsubsection{Local Geometry Similarity}
Geometric distortions, such as downsampling and geometry Gaussian noise, usually result in a change of the point density and point distribution, which is easily captured by the HVS. To quantitatively measure the point density and point distribution, we propose two statistical features within Euclidean distances between $\x_i$ and its neighbors in $N^{\X}_{\x_i,r}$ and $N^{\Y}_{\x_i,r}$ as follows:
\begin{equation}\label{eq:mean}
\mu^{\X}_{i,geo} = \frac{\sum_{\x_j\in N^{\X}_{\x_i,r}} \|\x_j^O-\x_i^O\|_2}{|N^{\X}_{\x_i,r}|} , 
\end{equation}

\begin{equation}\label{eq:var}
\sigma^{\X}_{i,geo} = \frac{\sum_{\x_j\in N^{\X}_{\x_i,r}}  (\|\x_j^O-\x_i^O\|_2 - \mu^{\X}_{i,geo})^2}{|N^{\X}_{\x_i,r}|}. 
\end{equation}
$\mu^{\Y}_{i,geo}$ and $\sigma^{\Y}_{i,geo}$ can be obtained via the same paradigm as Eq. (\ref{eq:mean}) and (\ref{eq:var}). Then we mimic the similarity measurement used in SSIM \cite{SSIM} to fuse the above statistical features as follows:
\begin{equation}
F_1=\frac{2\mu^{\X}_{i,geo}\cdot\mu^{\Y}_{i,geo}+T_1}{(\mu^{\X}_{i,geo})^2+(\mu^{\Y}_{i,geo})^2+T_1} \cdot \frac{2\sigma^{\X}_{i,geo}\cdot \sigma^{\Y}_{i,geo}+T_1}{(\sigma^{\X}_{i,geo})^2+(\sigma^{\Y}_{i,geo})^2+T_1},
\end{equation}
where $T_1$ are small constants to avoid instability. 
\subsubsection{Local Color Similarity}
It is widely believed that the HVS is more sensitive to contrast than absolute intensity \cite{Optimalcontrast}. Therefore, we try to calculate local color contrast in $N^{\X}_{\x_i,r}$ and $N^{\Y}_{\x_i,r}$ to reflect color distortions of PCs in our metric. Considering the sensitivity of the HVS to luminance channel \cite{GraphSIM}, we first calculate the luminance information according to the ITU-R Recommendation BT.709 \cite{YUVpara} as following, 
\begin{equation}
    \x_i^L = \x_i^C\cdot[0.257,0.504,0.098]^T+16.
\end{equation}

Similar to Eq. \eqref{eq:mean}, we then derive the $\mu^{\X}_{i,lum}$ by merging the luminance difference between $x_i$ and $N^{\X}_{\x_i,r}$ to represent the local color variation:

\begin{equation}\label{eq:meancolor}
\mu^{\X}_{i,lum} = \frac{\sum_{\x_j\in N^{\X}_{\x_i,r}} \|\x_j^{L}-\x_i^{L}\|_2}{|N^{\X}_{\x_i,r}|}.
\end{equation}
$\mu^{\Y}_{i,lum}$ can be obtained in the same way. Next, the quality feature reflecting color distortion is defined as
\begin{equation}
F_2=\frac{2\mu^{\X}_{i,lum}\cdot\mu^{\Y}_{i,lum}+T_1}{(\mu^{\X}_{i,lum})^2+(\mu^{\Y}_{i,lum})^2+T_1}.
\end{equation}
\subsubsection{Local Saliency Similarity}
According to Section \ref{sec:projection saliency}, we respectively calculate the saliency
maps of $\X$ and $\Y$ as $\S_{\X}=\{\x_i^S\}\in\R^N$ and $\S_{\Y}=\{\y_j^S\}\in\R^M$, respectively. Referring to some previous researches \cite{VSI}, human attention distribution changes along with different distortions, thus highly correlating with visual perception. Therefore, we can utilize the discrepancy between $\S_{\X}$ and $\S_{\Y}$ to assist in quality assessment. The mean of the saliency difference between $\x_i$ and $N^{\X}_{\x_i,r}$ is first obtained by
\begin{equation}\label{eq:meansal}
\mu^{\X}_{i,Sal} = \frac{\sum_{\x_j\in N^{\X}_{\x_i,r}} \|\x_j^{\S}-\x_i^{\S}\|_2}{|N^{\X}_{\x_i,r}|} , 
\end{equation}
$\mu^{\Y}_{i,Sal}$ can be calculated similarly. Then we define the quality feature reflecting the distance between two saliency maps as follows:
\begin{equation}
F_3=\frac{2\mu^{\X}_{i,sal}\cdot\mu^{\Y}_{i,sal}+T_2}{(\mu^{\X}_{i,sal})^2+(\mu^{\Y}_{i,sal})^2+T_2} ,
\end{equation}
where $T_2$ is another small constant.

\subsubsection{Global Similarity}
We have obtained three quality features for each point, which separately measure the local discrepancy of geometry structure, color contrast and saliency information. These features are then fused together to generate one local index:
\begin{equation}
SIM_{i}=F_1\cdot F_2\cdot F_3.
\end{equation}

\begin{table}[]
\centering
\caption{SROCC VALUES OBTAINED BY PQSM ON SJTU-PCQA DATABASE USING DIFFERENT 2D SALIENCY MODELS  }
\label{tab:2D_saliency_comparison}
\resizebox{88mm}{!}{
\renewcommand\arraystretch{1.2}
\begin{tabular}{cccccc}
\hline
      & Torra\cite{Torra}  & FES\cite{FES}    & UHF\cite{UHF}    & Murraymodel\cite{Murray} & SUN\cite{SUN}    \\ \hline
SROCC & 0.8745 & 0.8044 & 0.8838 & 0.8814      & 0.8308 \\ \hline
\end{tabular}}
\end{table}

An appropriate pooling strategy is desired to fuse all local indices. It has been widely accepted that not all points in point clouds share equal importance, and these points with high saliency values usually perform a more vital role in quality assessment. Therefore, the saliency map is naturally considered as the weighting factor for our pooling. Specifically, the final quality score is defined as follows:
\begin{equation}
Q =\frac{SIM_{i}\cdot \x_i^{\S}}{\sum_{i=1}^{N} \x_i^{\S}}.  
\end{equation}

\begin{table*}[pt]
\caption{PERFORMANCE COMPARISON OF DIFFERENT PCQA METRICS ON THE SJTU-PCQA, WPC, ICIP2020, AND M-PCCD DATASETS}
\label{tab:overall_performance}
\resizebox{181mm}{!}{
\setlength{\tabcolsep}{2.5pt}
\renewcommand\arraystretch{1.2}
\begin{tabular}{l|ccc|ccc|ccc|ccc|ccc}
\hline
{\color[HTML]{000000} Datasets} & \multicolumn{3}{c|}{{\color[HTML]{000000} SJTU-PCQA\cite{SJTUPCQA}}} & \multicolumn{3}{c|}{{\color[HTML]{000000} WPC\cite{WPC}}} & \multicolumn{3}{c|}{{\color[HTML]{000000} ICIP\cite{ICIP2020}}} & \multicolumn{3}{c|}{{\color[HTML]{000000} M-PCCD\cite{MPCCD}}} & \multicolumn{3}{c}{{\color[HTML]{000000} Weighted-AVE}} \\ \hline
{\color[HTML]{000000} Metrics} & {\color[HTML]{000000} PLCC} & {\color[HTML]{000000} SROCC} & {\color[HTML]{000000} RMSE} & {\color[HTML]{000000} PLCC} & {\color[HTML]{000000} SROCC} & {\color[HTML]{000000} RMSE} & {\color[HTML]{000000} PLCC} & {\color[HTML]{000000} SROCC} & {\color[HTML]{000000} RMSE} & {\color[HTML]{000000} PLCC} & {\color[HTML]{000000} SROCC} & {\color[HTML]{000000} RMSE} & {\color[HTML]{000000} PLCC} & {\color[HTML]{000000} SROCC} & {\color[HTML]{000000} RMSE} \\ \hline
{\color[HTML]{000000} p2point-MSE-PSNR\cite{mpeg}} & {\color[HTML]{000000} 0.477} & {\color[HTML]{000000} 0.577} & {\color[HTML]{000000} 2.133} & {\color[HTML]{000000} 0.578} & {\color[HTML]{000000} 0.566} & {\color[HTML]{000000} 18.70} & {\color[HTML]{000000} 0.951} & {\color[HTML]{000000} 0.935} & {\color[HTML]{000000} 0.351} & {\color[HTML]{000000} 0.853} & {\color[HTML]{000000} 0.868} & {\color[HTML]{000000} 0.710} & {\color[HTML]{000000} 0.619} & {\color[HTML]{000000} 0.641} & {\color[HTML]{000000} 10.30} \\
{\color[HTML]{000000} p2point-Hausdroff-PNSR\cite{mpeg}} & {\color[HTML]{000000} 0.498} & {\color[HTML]{000000} 0.527} & {\color[HTML]{000000} 2.105} & {\color[HTML]{000000} 0.403} & {\color[HTML]{000000} 0.258} & {\color[HTML]{000000} 20.97} & {\color[HTML]{000000} 0.593} & {\color[HTML]{000000} 0.532} & {\color[HTML]{000000} 0.914} & {\color[HTML]{000000} 0.609} & {\color[HTML]{000000} 0.370} & {\color[HTML]{000000} 1.079} & {\color[HTML]{000000} 0.473} & {\color[HTML]{000000} 0.364} & {\color[HTML]{000000} 11.56} \\
{\color[HTML]{000000} p2plane-MSE-PSNR\cite{mpeg}} & {\color[HTML]{000000} 0.511} & {\color[HTML]{000000} 0.573} & {\color[HTML]{000000} 2.086} & {\color[HTML]{000000} 0.487} & {\color[HTML]{000000} 0.446} & {\color[HTML]{000000} 20.02} & {\color[HTML]{000000} 0.951} & {\color[HTML]{000000} 0.945} & {\color[HTML]{000000} 0.350} & {\color[HTML]{000000} 0.871} & {\color[HTML]{000000} 0.885} & {\color[HTML]{000000} 0.667} & {\color[HTML]{000000} 0.584} & {\color[HTML]{000000} 0.581} & {\color[HTML]{000000} 10.96} \\
{\color[HTML]{000000} p2plane-Hausdroff-PNSR\cite{mpeg}} & {\color[HTML]{000000} 0.514} & {\color[HTML]{000000} 0.528} & {\color[HTML]{000000} 2.081} & {\color[HTML]{000000} 0.389} & {\color[HTML]{000000} 0.315} & {\color[HTML]{000000} 21.11} & {\color[HTML]{000000} 0.627} & {\color[HTML]{000000} 0.579} & {\color[HTML]{000000} 0.885} & {\color[HTML]{000000} 0.615} & {\color[HTML]{000000} 0.521} & {\color[HTML]{000000} 1.073} & {\color[HTML]{000000} 0.473} & {\color[HTML]{000000} 0.421} & {\color[HTML]{000000} 11.62} \\
{\color[HTML]{000000} PSNRyuv\cite{mpeg}} & {\color[HTML]{000000} 0.650} & {\color[HTML]{000000} 0.644} & {\color[HTML]{000000} 1.845} & {\color[HTML]{000000} 0.551} & {\color[HTML]{000000} 0.536} & {\color[HTML]{000000} 19.13} & {\color[HTML]{000000} 0.868} & {\color[HTML]{000000} 0.867} & {\color[HTML]{000000} 0.564} & {\color[HTML]{000000} 0.654} & {\color[HTML]{000000} 0.660} & {\color[HTML]{000000} 1.029} & {\color[HTML]{000000} 0.613} & {\color[HTML]{000000} 0.605} & {\color[HTML]{000000} 10.51} \\
{\color[HTML]{000000} GraphSIM\cite{GraphSIM}} & {\color[HTML]{000000} 0.856} & {\color[HTML]{000000} 0.541} & {\color[HTML]{000000} 1.253} & {\color[HTML]{000000} \textbf{0.694}} & {\color[HTML]{3166FF} \textbf{0.680}} & {\color[HTML]{000000} \textbf{16.49}} & {\color[HTML]{000000} 0.891} & {\color[HTML]{000000} 0.872} & {\color[HTML]{000000} 0.515} & {\color[HTML]{3166FF} \textbf{0.924}} & {\color[HTML]{FE0000} \textbf{0.939}} & {\color[HTML]{3166FF} \textbf{0.522}} & {\color[HTML]{000000} \textbf{0.786}} & {\color[HTML]{000000} 0.697} & {\color[HTML]{000000} \textbf{8.919}} \\
{\color[HTML]{000000} pointSSIM-geo\cite{pointSSIM}} & {\color[HTML]{000000} 0.754} & {\color[HTML]{000000} 0.699} & {\color[HTML]{000000} 1.593} & {\color[HTML]{000000} 0.440} & {\color[HTML]{000000} 0.341} & {\color[HTML]{000000} 20.58} & {\color[HTML]{000000} 0.896} & {\color[HTML]{000000} 0.905} & {\color[HTML]{000000} 0.505} & {\color[HTML]{000000} 0.831} & {\color[HTML]{000000} 0.834} & {\color[HTML]{000000} 0.757} & {\color[HTML]{000000} 0.614} & {\color[HTML]{000000} 0.550} & {\color[HTML]{000000} 11.15} \\
{\color[HTML]{000000} pointSSIM-col\cite{pointSSIM}} & {\color[HTML]{000000} 0.725} & {\color[HTML]{000000} 0.704} & {\color[HTML]{000000} 1.672} & {\color[HTML]{000000} 0.510} & {\color[HTML]{000000} 0.454} & {\color[HTML]{000000} 19.71} & {\color[HTML]{000000} 0.904} & {\color[HTML]{000000} 0.865} & {\color[HTML]{000000} 0.486} & {\color[HTML]{FE0000} \textbf{0.926}} & {\color[HTML]{3166FF} \textbf{0.918}} & {\color[HTML]{FE0000} \textbf{0.514}} & {\color[HTML]{000000} 0.658} & {\color[HTML]{000000} 0.620} & {\color[HTML]{000000} 10.68} \\
{\color[HTML]{000000} MPED-1-norm\cite{MPED}} & {\color[HTML]{FE0000} \textbf{0.898}} & {\color[HTML]{3166FF} \textbf{0.886}} & {\color[HTML]{FE0000} \textbf{1.067}} & {\color[HTML]{3166FF} \textbf{0.695}} & {\color[HTML]{000000} \textbf{0.673}} & {\color[HTML]{3166FF} \textbf{16.47}} & {\color[HTML]{3166FF} \textbf{0.964}} & {\color[HTML]{3166FF} \textbf{0.951}} & {\color[HTML]{3166FF} \textbf{0.303}} & {\color[HTML]{000000} 0.846} & {\color[HTML]{000000} 0.869} & {\color[HTML]{000000} 0.726} & {\color[HTML]{3166FF} \textbf{0.790}} & {\color[HTML]{FE0000} \textbf{0.778}} & {\color[HTML]{3166FF} \textbf{8.883}} \\
{\color[HTML]{000000} MPED-2-norm\cite{MPED}} & {\color[HTML]{FE0000} \textbf{0.898}} & {\color[HTML]{FE0000} \textbf{0.890}} & {\color[HTML]{3166FF} \textbf{1.070}} & {\color[HTML]{000000} 0.637} & {\color[HTML]{000000} 0.624} & {\color[HTML]{000000} 17.67} & {\color[HTML]{FE0000} \textbf{0.967}} & {\color[HTML]{FE0000} \textbf{0.961}} & {\color[HTML]{FE0000} \textbf{0.290}} & {\color[HTML]{000000} 0.867} & {\color[HTML]{000000} 0.887} & {\color[HTML]{000000} 0.678} & {\color[HTML]{000000} 0.763} & {\color[HTML]{000000} \textbf{0.757}} & {\color[HTML]{000000} 9.493} \\ \hline
{\color[HTML]{000000} PQSM (proposed)} & {\color[HTML]{000000} \textbf{0.894}} & {\color[HTML]{000000} \textbf{0.884}} & {\color[HTML]{000000} \textbf{1.086}} & {\color[HTML]{FE0000} \textbf{0.753}} & {\color[HTML]{FE0000} \textbf{0.737}} & {\color[HTML]{FE0000} \textbf{15.08}} & {\color[HTML]{000000} 0.909} & {\color[HTML]{000000} 0.902} & {\color[HTML]{000000} 0.474} & {\color[HTML]{000000} \textbf{0.899}} & {\color[HTML]{000000} \textbf{0.911}} & {\color[HTML]{000000} \textbf{0.597}} & {\color[HTML]{FE0000} \textbf{0.823}} & {\color[HTML]{FE0000} \textbf{0.814}} & {\color[HTML]{FE0000} \textbf{8.160}} \\ \hline
\end{tabular}}
\end{table*}

\section{EXPERIMENTS}
In this section, we present the performance of our PQSM and other state-of-art PCQA metrics on four publicly accessible databases and investigate the effectiveness of the induced saliency information.
\subsection{Database and Evaluation Criteria}
\begin{itemize}
\item[$\bullet$]
SJTU-PCQA: It has 9 reference PCs and 378 distorted samples. Each reference PC is impaired by 7 different types of distortion under 6 levels, including individual distortions and superimposed distortions. 
\item[$\bullet$]
WPC: It contains 20 reference PCs and 740 distorted samples. 5 types of distortion are exerted on each reference PC, including downsampling, Gaussian noise contamination, G-PCC(Octree), G-PCC(Trisoup) and V-PCC. 
\item[$\bullet$]
ICIP2020: It contains 6 reference PCs and 90 distorted samples. The distortion results from G-PCC and V-PCC, and each is distributed in 5 levels.
\item[$\bullet$]
M-PCCD: It contains 8 reference PCs and 232 distorted samples resulting from MPEG-encoder. The distortions include V-PCC and four variations of G-PCC.
\end{itemize}

The Pearson linear correlation coefficient (PLCC), the Spearman rank order correlation coefficient (SROCC), and the root mean square error (RMSE) are utilized to evaluate the performance of different models. The logistic regression recommended by VQEG \cite{video2003final} is used to map the dynamic range
of the scores from objective quality assessment models into a common score.

\subsection{Selection of Model Parameters and 2D Saliency Method}
Here we present the detailed parameter setting and the selection of the 2D saliency method used in Section \ref{sec:projection saliency} .

\textbf{$\sigma_s$ in the weight calculation.} We determine $\sigma_s$ in Eq. \eqref{eq:depth saliency} as one tenth of the maximum side length of the bounding box, alleviating the impacts from different scales of PCs.

\textbf{$r$ in the neighborhood construction} We set $r$ as the average of distances between each point in the reference PC and the $10$-th nearest point in the distorted PC.

\textbf{$T_1$ and $T_2$ in the similarity pooling.} $T_1$ is set to be 0.001. $T_2$ is set to be $10^{-14}$ due to the small weight values after the process of normalization. 

\textbf{2D saliency method.} For the 2D saliency detection method used in Section \ref{sec:projection saliency}, we test several eminent algorithms on SJTU-PCQA and show the performance in terms of SROCC in TABLE \ref{tab:2D_saliency_comparison}. We can see that UHF provides the best performance, which is consequently utilized in our metric.

\subsection{Overall Performance Comparison}
We evaluate our proposed method and several other state-of-art PCQA metrics on four databases and list the performance in TABLE \ref{tab:overall_performance}. The top three performances of each database are indicated in bold with red, blue and black, respectively.

According to TABLE \ref{tab:overall_performance}, we can see that the proposed PQSM achieves competitive performance across the four databases. Specifically, PQSM is among the top three performances on three databases, i.e., SJTU-PCQA, WPC and M-PCCD. In particular, PQSM achieves the best correlation on WPC containing the most samples, providing (PLCC, SROCC, RMSE)=(0.753, 0.737, 15.08). Moreover, to evaluate the performance of PCQA across multiple databases, we illustrate the weighted average criteria based on the size of four databases in the last column of TABLE \ref{tab:overall_performance}. 
We can see that the proposed PQSM achieves the highest consistency in terms of weighted average performance, presenting (PLCC, SROCC, RMSE)=(0.823, 0.814, 8.160). It further demonstrates the effectiveness and robustness of our method.


\begin{table}[]
\caption{PERFORMANCE OF FEATURES OF PQSM ON THE SJTU-PCQA DATABASE}
\centering
\label{tab:feature_comparison}
\resizebox{88mm}{!}{
\renewcommand\arraystretch{1.2}
\begin{tabular}{c|c|ccc}
\hline
feature & pooling & PLCC & SROCC & RMSE \\ \hline
$F_1$ & AVE & 0.8405 & 0.7708 & 1.314 \\
$F_2$ & AVE & 0.8326 & 0.8305 & 1.343 \\
$F_3$ & AVE & 0.6205 & 0.5811 & 1.903 \\
$F1+F2$ & AVE & 0.8888 & 0.8790 & 1.111 \\
$F1+F2+F3$ & AVE & 0.8911 & 0.8824 & 1.101 \\ \hline
$F1+F2+F3$ & SAW & 0.8941 & 0.8838 & 1.086 \\ \hline
\end{tabular}}
\end{table}

\subsection{Ablation Study}
To investigate the impact of saliency information in PQSM, we measure the performances resulting from different feature combinations with different pooling strategies on SJTU-PCQA, which are listed in TABLE \ref{tab:feature_comparison}. Note that $AVE$ and $SAW$ represent average pooling and saliency-based weighting, respectively. From TABLE \ref{tab:feature_comparison}, we can see that both saliency feature and saliency-based weighting improve the model performance, which states that introducing proper saliency information is beneficial for PCQA tasks.
\section{CONCLUSION}
In this letter, we propose a new FR-PCQA metric involving visual saliency information. Inspired by the success of visual saliency in IQA, we make the first attempt to apply it to PCQA. Specifically, we introduce depth information to induce the intrinsic geometric characteristics of PCs into the projection-based saliency maps. Then we propose three quality features to separately measure local discrepancies of geometry structure, color contrast and saliency
information. A saliency-based pooling strategy is adopted to generate the final quality score. Our method is examined by extensive experiments on four publicly accessible PCQA databases and shows competitive performance compared to the state-of-the-art metrics.


\clearpage
\bibliographystyle{IEEEtran}
\bibliography{IEEEabrv,reference}

\begin{thebibliography}{10}
\providecommand{\url}[1]{#1}
\csname url@samestyle\endcsname
\providecommand{\newblock}{\relax}
\providecommand{\bibinfo}[2]{#2}
\providecommand{\BIBentrySTDinterwordspacing}{\spaceskip=0pt\relax}
\providecommand{\BIBentryALTinterwordstretchfactor}{4}
\providecommand{\BIBentryALTinterwordspacing}{\spaceskip=\fontdimen2\font plus
\BIBentryALTinterwordstretchfactor\fontdimen3\font minus
  \fontdimen4\font\relax}
\providecommand{\BIBforeignlanguage}[2]{{%
\expandafter\ifx\csname l@#1\endcsname\relax
\typeout{** WARNING: IEEEtran.bst: No hyphenation pattern has been}%
\typeout{** loaded for the language `#1'. Using the pattern for}%
\typeout{** the default language instead.}%
\else
\language=\csname l@#1\endcsname
\fi
#2}}
\providecommand{\BIBdecl}{\relax}
\BIBdecl

\bibitem{Navigation}
M.~Whitty, S.~Cossell, K.~S. Dang, J.~Guivant, and J.~Katupitiya, ``{Autonomous
  Navigation Using A Real-time 3D Point Cloud},'' in \emph{Australas. Conf.
  Robot. Automat.}, 2010, pp. 1--3.

\bibitem{robotics}
M.~Liu, ``Robotic online path planning on point cloud,'' \emph{IEEE Trans.
  Cybern.}, vol.~46, no.~5, pp. 1217--1228, 2016.

\bibitem{mixedreal}
F.~Capraro and S.~Milani, ``Rendering-aware point cloud coding for mixed
  reality devices,'' in \emph{2019 IEEE Int. Conf. Image Process.}, 2019, pp.
  3706--3710.

\bibitem{Compre1}
S.~Schwarz, M.~Preda, V.~Baroncini, M.~Budagavi, P.~Cesar, P.~A. Chou, R.~A.
  Cohen, M.~Krivokuća, S.~Lasserre, Z.~Li, J.~Llach, K.~Mammou, R.~Mekuria,
  O.~Nakagami, E.~Siahaan, A.~Tabatabai, A.~M. Tourapis, and V.~Zakharchenko,
  ``{Emerging MPEG Standards for Point Cloud Compression},'' \emph{IEEE J.
  Emerg. Sel. Topics Circuits Syst.}, vol.~9, no.~1, pp. 133--148, Dec. 2019.

\bibitem{Compre2}
R.~Schnabel and R.~Klein, ``{Octree-based Point-Cloud Compression.}'' in
  \emph{Symp. on Point-Based Graph.}\hskip 1em plus 0.5em minus 0.4em\relax The
  Eurographics Association, 2006, pp. 111--120.

\bibitem{transmiss}
G.-R. Jang, Y.-D. Shin, J.-H. Park, and M.-H. Baeg, ``{Real-time Point-Cloud
  Data Transmission for Teleoperation Using H.264/AVC},'' in \emph{IEEE Int.
  Symp. Saf. Secur. Rescue Robot.}, Oct. 2014, pp. 1--6.

\bibitem{SJTUPCQA}
Q.~Yang, H.~Chen, Z.~Ma, Y.~Xu, R.~Tang, and J.~Sun, ``{Predicting the
  Perceptual Quality of Point Cloud: A 3D-to-2D Projection-Based
  Exploration},'' \emph{IEEE Trans. Multimedia}, vol.~23, pp. 3877--3891, Oct.
  2021.

\bibitem{p2po}
R.~Mekuria, K.~Blom, and P.~Cesar, ``{Design, Implementation, and Evaluation of
  a Point Cloud Codec for Tele-Immersive Video},'' \emph{IEEE Trans. Circuits
  Syst. for Video Technology}, vol.~27, no.~4, pp. 828--842, Mar. 2017.

\bibitem{p2pl}
D.~Tian, H.~Ochimizu, C.~Feng, R.~Cohen, and A.~Vetro, ``{Geometric Distortion
  Metrics for Point Cloud Compression},'' in \emph{IEEE Int. Conf. Image
  Process.}, Sept. 2017, pp. 3460--3464.

\bibitem{pointSSIM}
E.~Alexiou and T.~Ebrahimi, ``Towards a point cloud structural similarity
  metric,'' in \emph{IEEE Int. Conf. Multimedia Expo. Workshops}, 2020, pp.
  1--6.

\bibitem{GraphSIM}
Q.~Yang, Z.~Ma, Y.~Xu, Z.~Li, and J.~Sun, ``{Inferring Point Cloud Quality via
  Graph Similarity},'' \emph{IEEE Trans. Pattern Anal. Mach. Intell.}, vol.~44,
  no.~6, pp. 3015--3029, Dec. 2022.

\bibitem{MSGraphSIM}
Y.~Zhang, Q.~Yang, and Y.~Xu, \emph{{MS-GraphSIM: Inferring Point Cloud Quality
  via Multiscale Graph Similarity}}.\hskip 1em plus 0.5em minus 0.4em\relax NY,
  USA: Association for Computing Machinery, Oct. 2021, p. 1230–1238.

\bibitem{MPED}
Q.~{Yang}, Y.~{Zhang}, S.~{Chen}, Y.~{Xu}, J.~{Sun}, and Z.~{Ma}, ``{MPED:
  Quantifying Point Cloud Distortion based on Multiscale Potential Energy
  Discrepancy},'' \emph{arXiv e-prints}, p. arXiv:2103.02850, Mar. 2021.

\bibitem{VSI}
L.~Zhang, Y.~Shen, and H.~Li, ``{VSI: A Visual Saliency-Induced Index for
  Perceptual Image Quality Assessment},'' \emph{IEEE Trans. Image Process.},
  vol.~23, no.~10, pp. 4270--4281, Aug. 2014.

\bibitem{SalGuo}
Y.~Guo, F.~Wang, and J.~Xin, ``{Point-wise Saliency Detection on 3D Point
  Clouds via Covariance Descriptors},'' \emph{Vis. Comput.}, vol.~34, no.~10,
  pp. 1325--1338, 2018.

\bibitem{SalDing}
X.~Ding, W.~Lin, Z.~Chen, and X.~Zhang, ``{Point Cloud Saliency Detection by
  Local and Global Feature Fusion},'' \emph{IEEE Trans. Image Process.},
  vol.~28, no.~11, pp. 5379--5393, May 2019.

\bibitem{SalMaps}
V.~F. Figueiredo, G.~L. Sandri, R.~L. de~Queiroz, and P.~A. Chou, ``{Saliency
  Maps for Point Clouds},'' in \emph{2020 IEEE 22nd International Workshop on
  Multimedia Signal Processing (MMSP)}, Sept. 2020, pp. 1--5.

\bibitem{DepthMatters}
C.~Lang, T.~V. Nguyen, H.~Katti, K.~Yadati, M.~Kankanhalli, and S.~Yan,
  ``{Depth Matters: Influence of Depth Cues on Visual Saliency},'' in
  \emph{European Conf. on Comput. Vision}.\hskip 1em plus 0.5em minus
  0.4em\relax Springer, 2012, pp. 101--115.

\bibitem{UHF}
H.~R.~Tavakoli and J.~Laaksonen, ``{Bottom-Up Fixation Prediction Using
  Unsupervised Hierarchical Models},'' in \emph{Comput. Vision -- ACCV 2016
  Workshops}, C.-S. Chen, J.~Lu, and K.-K. Ma, Eds.\hskip 1em plus 0.5em minus
  0.4em\relax Cham: Springer International Publishing, 2017, pp. 287--302.

\bibitem{SSIM}
Z.~Wang, A.~Bovik, H.~Sheikh, and E.~Simoncelli, ``Image quality assessment:
  from error visibility to structural similarity,'' \emph{IEEE Trans. Imag.
  Process.}, vol.~13, no.~4, pp. 600--612, 2004.

\bibitem{Optimalcontrast}
\BIBentryALTinterwordspacing
X.~Qian, J.~Han, G.~Cheng, and L.~Guo, ``Optimal contrast based saliency
  detection,'' \emph{Pattern Recognit. Lett.}, vol.~34, no.~11, pp. 1270--1278,
  2013. [Online]. Available:
  \url{https://www.sciencedirect.com/science/article/pii/S0167865513001591}
\BIBentrySTDinterwordspacing

\bibitem{YUVpara}
I.~ITU, ``{Parameter Values for The HDTV Standards for Production and
  International Programme Exchange},'' \emph{Recommendation ITU-R BT}, pp.
  709--5, 2002.

\bibitem{Torra}
A.~Torralba, A.~Oliva, M.~S. Castelhano, and J.~M. Henderson, ``{Contextual
  Guidance of Eye Movements and Attention in Real-world Scenes: The Role of
  Global Features in Object Search.}'' \emph{Psychological Rev.}, vol. 113,
  no.~4, p. 766, 2006.

\bibitem{FES}
H.~Rezazadegan~Tavakoli, E.~Rahtu, and J.~Heikkil{\"a}, ``{Fast and Efficient
  Saliency Detection Using Sparse Sampling and Kernel Density Estimation},'' in
  \emph{Scandinavian Conf. on Image Anal.}\hskip 1em plus 0.5em minus
  0.4em\relax Springer, 2011, pp. 666--675.

\bibitem{Murray}
N.~Murray, M.~Vanrell, X.~Otazu, and C.~A. Parraga, ``{Saliency Estimation
  Using A Non-parametric Low-level Vision Model},'' in \emph{Conf. on Comput.
  Vision and Pattern Recognit.}, June 2011, pp. 433--440.

\bibitem{SUN}
L.~Zhang, M.~H. Tong, T.~K. Marks, H.~Shan, and G.~W. Cottrell, ``{SUN: A
  Bayesian Framework for Saliency Using Natural Statistics},'' \emph{J.
  Vision}, vol.~8, no.~7, pp. 32--32, 12 2008.

\bibitem{WPC}
Q.~Liu, H.~Su, Z.~Duanmu, W.~Liu, and Z.~Wang, ``{Perceptual Quality Assessment
  of Colored 3D Point Clouds},'' \emph{IEEE Trans. Vis. Comput. Graphics}, pp.
  1--1, Apr. 2022.

\bibitem{ICIP2020}
S.~Perry, H.~P. Cong, L.~A. da~Silva~Cruz, J.~Prazeres, M.~Pereira,
  A.~Pinheiro, E.~Dumic, E.~Alexiou, and T.~Ebrahimi, ``Quality evaluation of
  static point clouds encoded using mpeg codecs,'' in \emph{2020 IEEE Int.
  Conf. on Image Process.}, 2020, pp. 3428--3432.

\bibitem{MPCCD}
E.~Alexiou, I.~Viola, T.~M. Borges, T.~A. Fonseca, R.~L. de~Queiroz, and
  T.~Ebrahimi, ``A comprehensive study of the rate-distortion performance in
  mpeg point cloud compression,'' \emph{APSIPA Trans. Signal Inf. Process.},
  vol.~8, p. e27, 2019.

\bibitem{mpeg}
\BIBentryALTinterwordspacing
Mpeg reference software. [Online]. Available:
  \url{http://mpegx.int-evry.fr/software/MPEG/PCC/TM/mpeg-pcc-dmetric}
\BIBentrySTDinterwordspacing

\bibitem{video2003final}
VEQG, ``Final report from the video quality experts group on the validation of
  objective models of video quality assessment,'' \emph{[online]. Availabel:
  http://www.its.bldrdoc.gov/vqeg/vqeg-home.aspx}.

\end{thebibliography}

\end{document}